\documentclass[conference]{IEEEtran}
\IEEEoverridecommandlockouts
\usepackage{cite}
\usepackage{amsmath,amssymb,amsfonts}
\usepackage{algorithmic}
\usepackage{graphicx}
\usepackage{textcomp}
\usepackage{xcolor}

\usepackage{epsfig}
\usepackage{times}
\usepackage{epsfig}
\usepackage{graphicx}
\usepackage{amsmath}
\usepackage{amssymb}
\usepackage{cite}

\usepackage{color}
\usepackage {ulem}
\usepackage{bm}
\usepackage{algorithm}
\usepackage{algorithmic}
\usepackage{multirow}
\usepackage{subfigure}
\usepackage{amsthm}
\usepackage{booktabs}
\usepackage{textcomp}
\usepackage{mathrsfs}
\usepackage{makecell}
\usepackage{diagbox}
\usepackage{setspace}

\usepackage{bbding}
\usepackage{pifont}
\usepackage{wasysym}

\usepackage{epstopdf}

\usepackage{hyperref}

\def\B{\mathbf{B}}
\def\S{\mathbf{S}}

\def\R{\mathbf{R}}

\def\X{\mathbf{X}}
\def\Y{\mathbf{Y}}
\def\F{\mathcal{F}}

\def\BibTeX{{\rm B\kern-.05em{\sc i\kern-.025em b}\kern-.08em
    T\kern-.1667em\lower.7ex\hbox{E}\kern-.125emX}}

\begin{document}

\title{Towards General and Fast Video Derain via Knowledge Distillation\\}

\author{\IEEEauthorblockN{Defang Cai}
\IEEEauthorblockA{\textit{College of Computer Science \& Technology} \\
\textit{Zhejiang University of Technology }\\
Hangzhou, China \\
2112112198@zjut.edu.cn}
\and
\IEEEauthorblockN{Pan Mu*\thanks{* Corresponding author.}}
\IEEEauthorblockA{\textit{College of Computer Science \& Technology} \\
\textit{Zhejiang University of Technology}\\
Hangzhou, China \\
panmu@zjut.edu.cn}
\and
\IEEEauthorblockN{Sixian Chan}
\IEEEauthorblockA{\textit{College of Computer Science \& Technology} \\
\textit{Zhejiang University of Technology}\\
Hangzhou, China \\
sxchan@zjut.edu.cn}
\and
\IEEEauthorblockN{Zhanpeng Shao}
\IEEEauthorblockA{\textit{College of Information Science} \\ \textit{and Engineering} \\
\textit{Hunan Normal University}\\
Hunan, China \\
zpshao@hunnu.edu.cn}
\and
\IEEEauthorblockN{Cong Bai }
\IEEEauthorblockA{\textit{College of Computer Science \& Technology} \\
\textit{ Zhejiang University of Technology }\\
Hangzhou, China \\
congbai@zjut.edu.cn}
}

\maketitle

\begin{abstract}
As a common natural weather condition, rain can obscure video frames and thus affect the performance of the visual system, so video derain receives a lot of attention. In natural environments, rain has a wide variety of streak types, which increases the difficulty of the rain removal task. In this paper, we propose a Rain Review-based General video derain Network via knowledge distillation (named RRGNet) that handles different rain streak types with one pre-training weight. Specifically, we design a frame grouping-based encoder-decoder network that makes full use of the temporal information of the video. Further, we use the old task model to guide the current  model in learning new rain streak types while avoiding forgetting. To consolidate the network's ability to derain, we design a rain review module to play back data from old tasks for the current model. The experimental results show that our developed general method achieves the best results in terms of running speed and derain effect. 
\end{abstract}


\section{Introduction}
Rain is one of the most common types of natural weather. In rainy conditions, rainfall can block the photographic equipment from capturing the background environment and affect the subsequent visual technique. Removing the effects of rain on photography has become a widely researched problem, and there are already many effective deep learning methods. However, these methods require different pre-training weights for different rain streak types to work optimally, making it challenging to apply them to real-world environments.
\begin{figure}[t]
	\centering 
	\begin{center}
		\begin{tabular}{c@{\extracolsep{0.2em}}c}
			
			\includegraphics[width=0.23\textwidth]{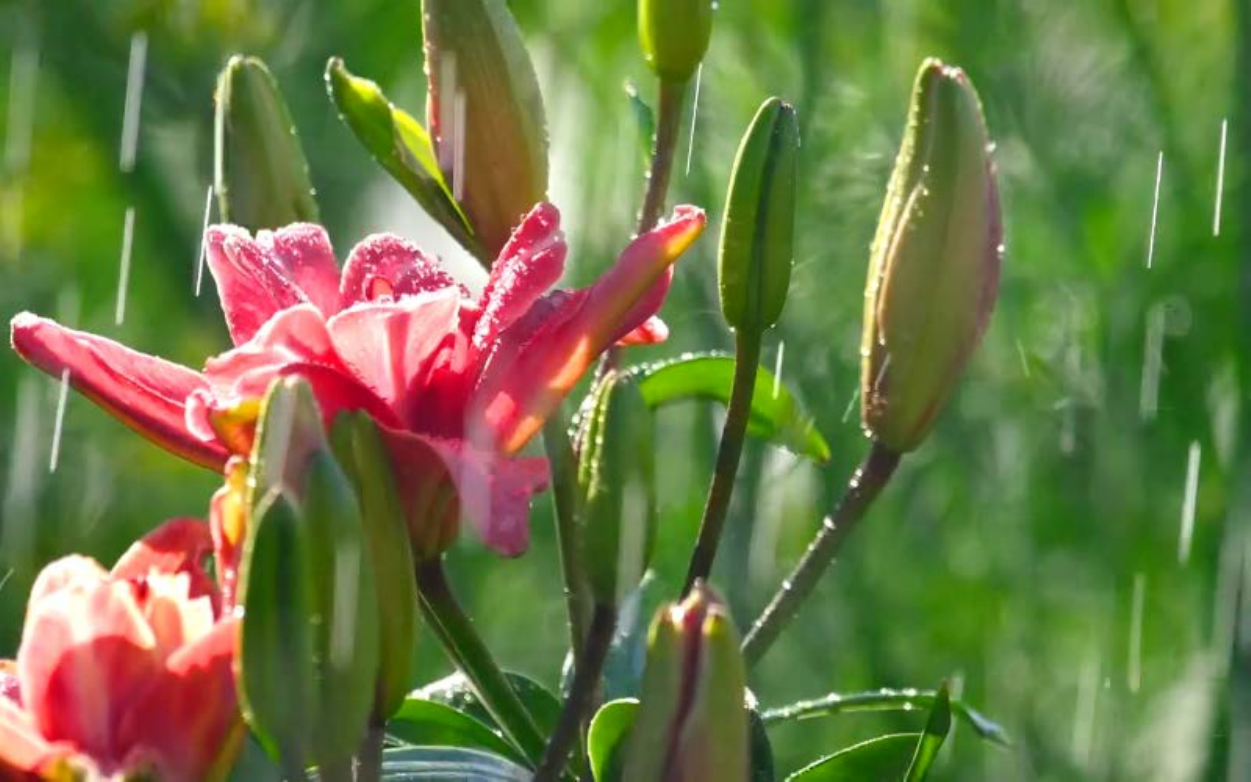}&
			\includegraphics[width=0.23\textwidth]{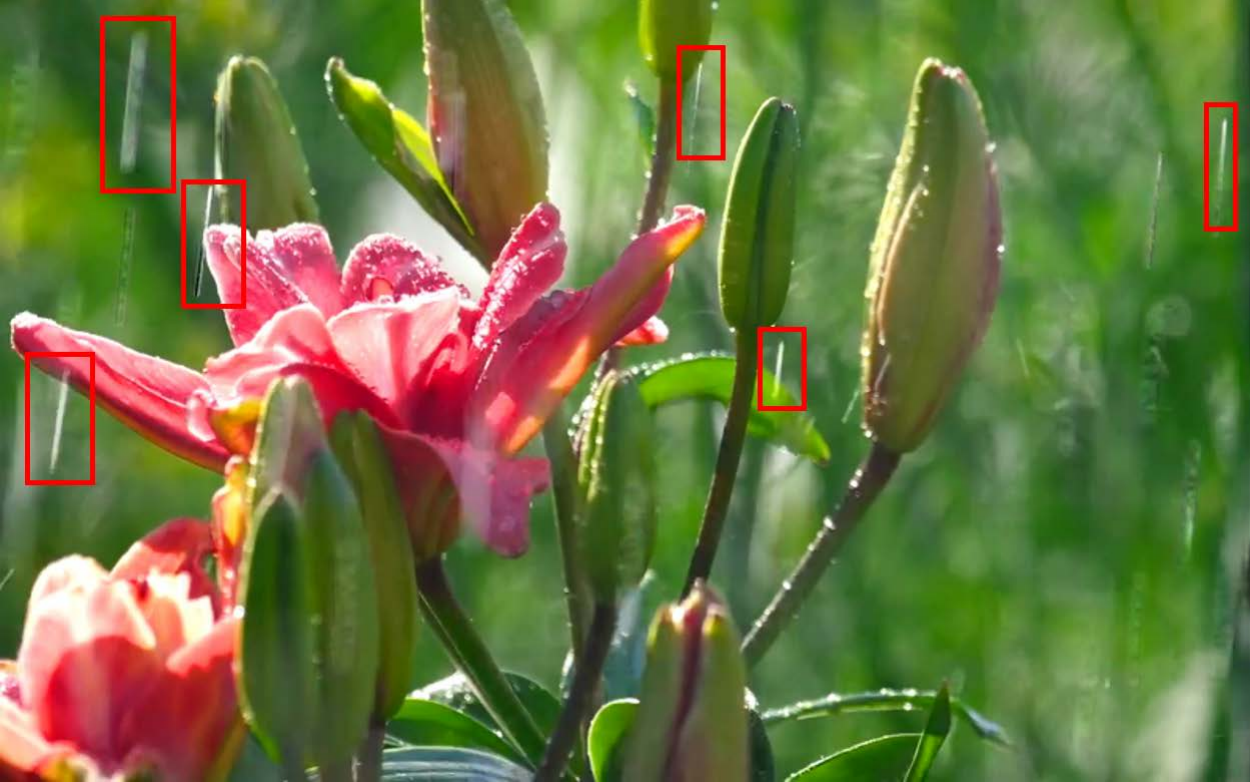}\\
			\footnotesize Input &\footnotesize ESTINet~\cite{zhang2022enhanced}  \\
			
			\includegraphics[width=0.23\textwidth]{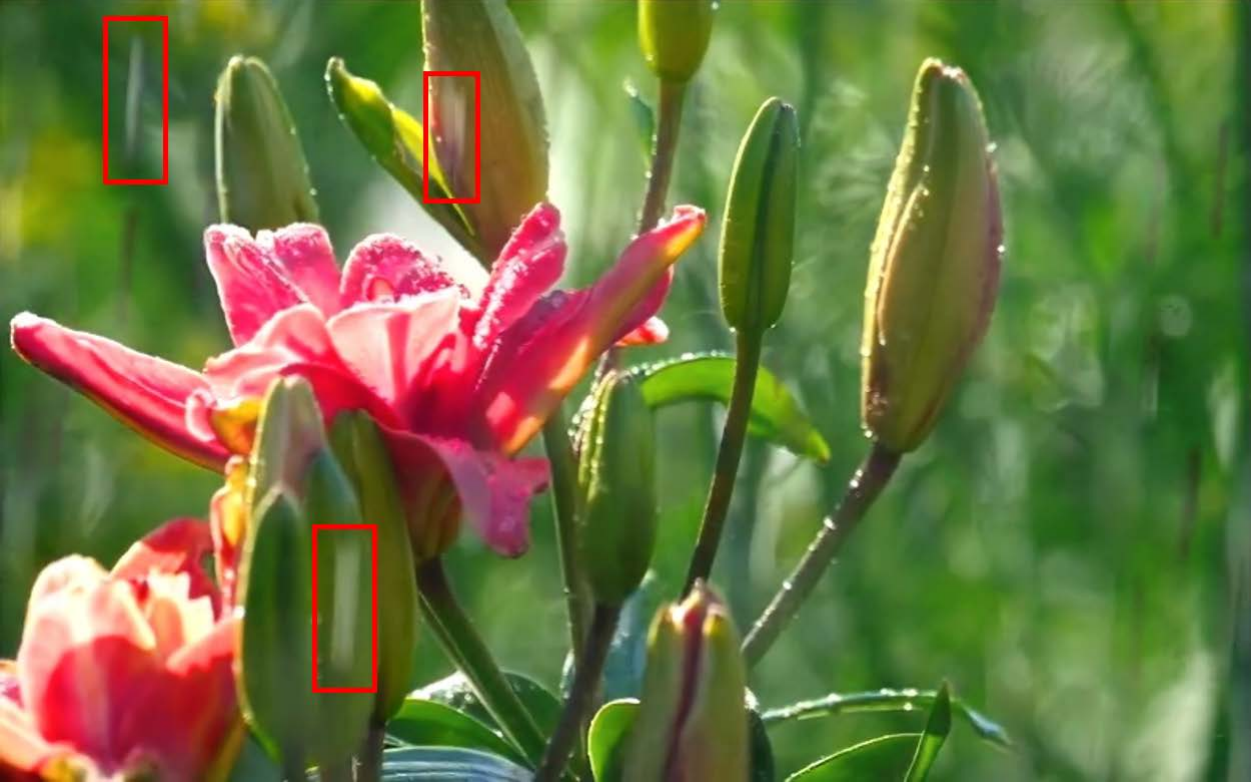}&
			\includegraphics[width=0.23\textwidth]{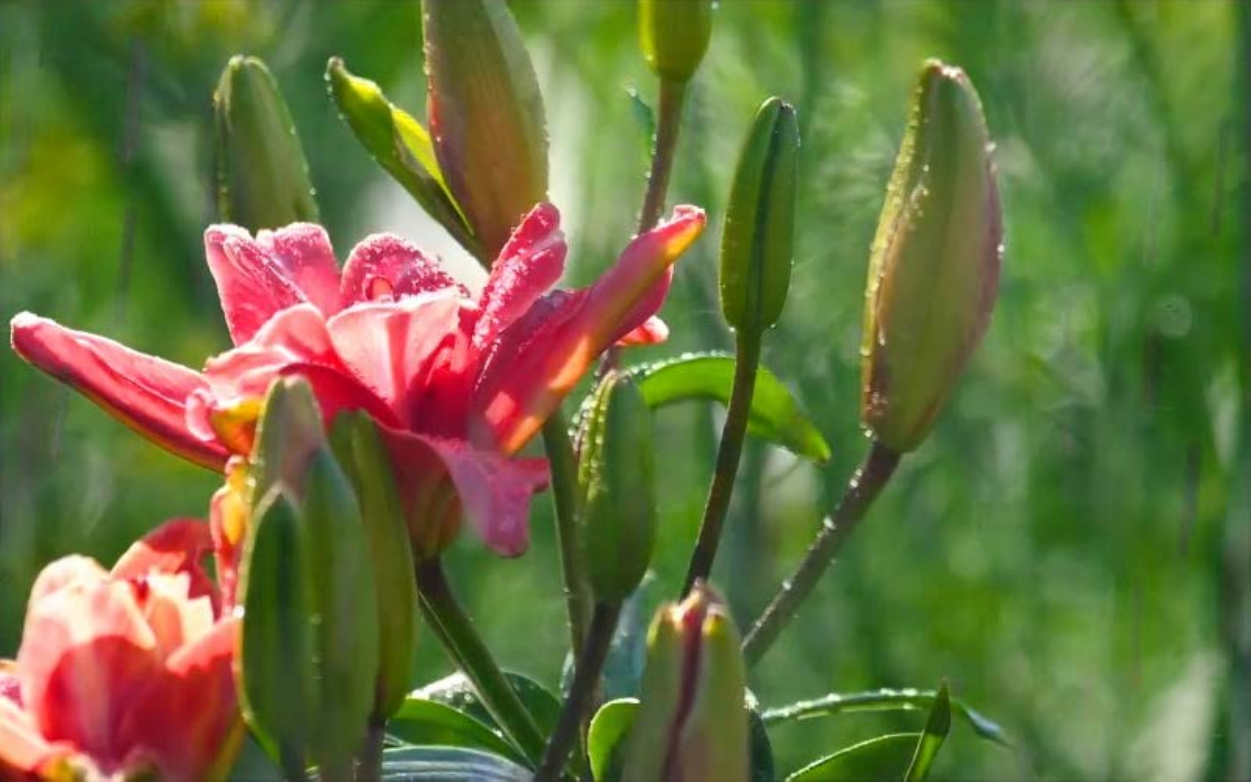}\\
			
			\footnotesize RMFD~\cite{yang2021recurrent} & \footnotesize Ours 
		\end{tabular}
	\end{center}
	\vspace{-3mm}
	\caption{The proposed method compares with various SOTA video derain methods on the real-word video. It is clear that our method achieves the best rain removal results.}\label{fig:first} 
\end{figure}

\begin{figure*}[htb!]
	\centering 
	\begin{center}
		\begin{tabular}{c@{\extracolsep{0.2em}}}
			\includegraphics[width=0.93\textwidth]{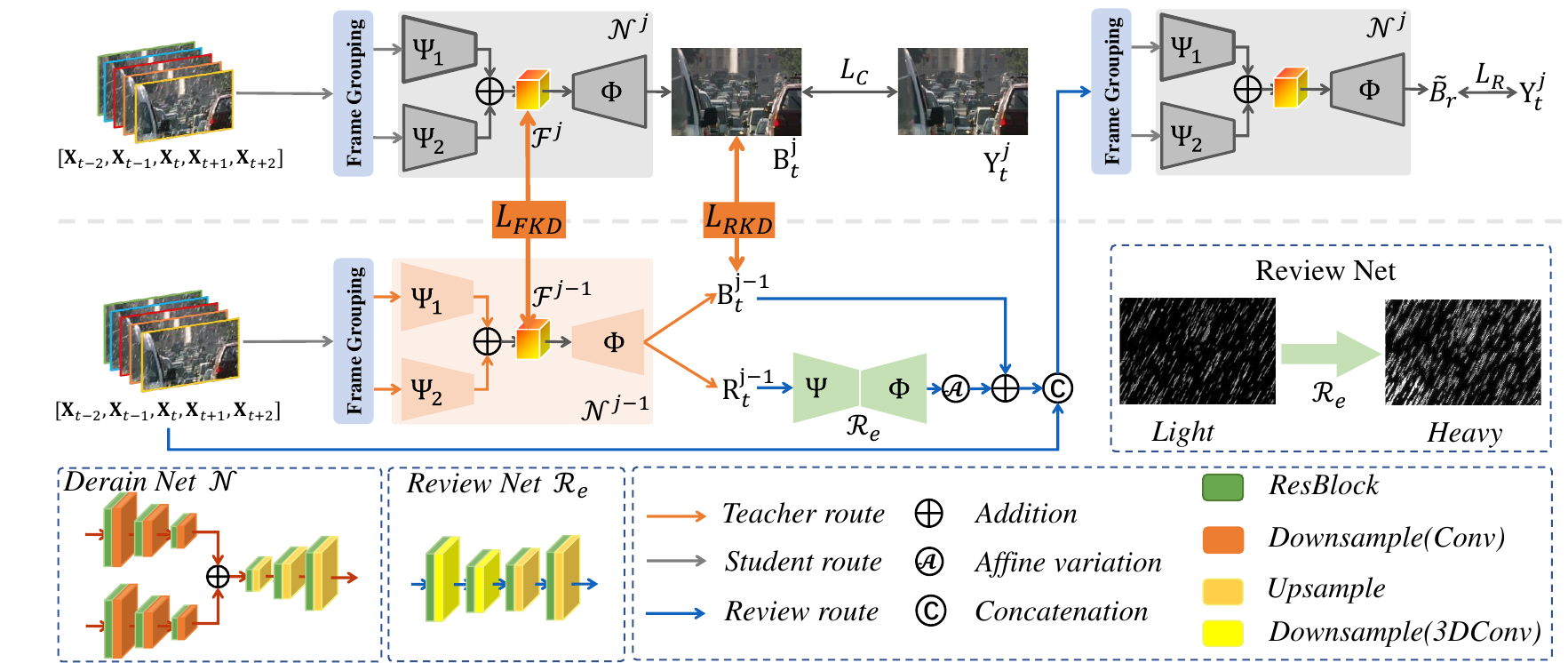}
		\end{tabular}
	\end{center}
	\vspace{-3mm}
	\caption{Training flow chart of our method. It should be noted that during the evaluation phase the derain network $\mathcal{N}^j$ does all the rain removal work alone.}\label{fig:flow chart} 
\end{figure*}

The derain task is divided into image derain and video derain depending on the input. Recently, many effective approaches are proposed in the field of image derain, such as Adaptive filtering~\cite{kim2013single}, sparse coding~\cite{chen2014visual}, dictionary learning~\cite{luo2015removing}, data-driven based methods~\cite{liu2020investigating,mu2018learning}, and deep learning methods~\cite{wang2021multi,zamir2021multi,wang2021rain,fu2021rain,chen2022unpaired,zamir2022restormer}. Compared to image derain, video derain requires better use of temporal information, which poses challenges. Some early video derain studies attempts to complete derain tasks based on physical priors. Such as some frequency domain based method~\cite{barnum2007spatio}, low-rank structure method~\cite{kim2015video,chen2013generalized}, sparse matrix method~\cite{li2018video,ren2017video}, the blurred Gaussian model~\cite{wei2017should}, tensor structures~\cite{jiang2017novel,dong2016color}, and some rain streak priors~\cite{tripathi2012video}. All these traditional methods require input to satisfy certain physical assumptions and they are difficult to process complex natural rain streaks.

In recent years, along with the development of deep learning, data-based methods are receiving attention from many researchers. The most common deep learning derain method is to separate the rain layers using convolution, e.g.~\cite{zhang2022enhanced,mu2021triple}. where Zhang $et\,al.$~\cite{zhang2022enhanced} takes advantage of deep residual networks and LSTM convolution to effectively combine spatial and temporal features. Mu $et\,al.$~\cite{mu2021triple} proposes a dynamic model based on the NAS search structure to address the shortcomings of the CNN approach. To make better use of the timing information in the video, some methods align the video frames before derain, e.g.~\cite{yang2019frame,yan2021self,su2022complex,yan2022feature}. where Su $et\,al.$~\cite{su2022complex} and Yang $et\,al.$~\cite{yang2019frame} complete the frame alignment using the optical flow method. And Yan $et\,al.$~\cite{yan2021self} uses deformable convolution instead of the optical flow method because of the instability of light in rainy conditions. Some researchers try to combine deep learning methods and model-driven methods, such as~\cite{liu2018erase,chen2018robust,yang2021recurrent}. Yang $et\,al.$~\cite{yang2021recurrent} combines adversarial learning and physical priors to design a two-stage progressive network to handle rain accumulation.

However, none of these methods can handle all types of rain streaks using one pre-trained model. This is because all of the above deep learning methods need to address the problem of catastrophic forgetting. The catastrophic forgetting problem means that after the model is already partially memorizing rain streak knowledge when it learns a completely new type of rain streak, the model immediately forgets how the previous derain task is handled. This makes it necessary for each model to prepare different pre-training weights for different rain streaks. This can greatly affect the application of derain networks in reality.

To mitigate this problem, we develop a Rain Review-based General video derain Network via knowledge distillation (named RRGNet) that handles different rain streak types with one pre-training weight. To make better use of the temporal information of the video, we propose a frame grouping based encoder-decoder network which can extract rain streaks information from different frame rates. And we design a feature and response distillation module which effectively preserves the model's memory of old tasks. To further remember past knowledge, we design a rain review module to generate rain streaks of old tasks. These rain streaks help models review knowledge of old tasks. Our contribution is summarised as follows:

\begin{itemize}
	\item We propose a general video derain network via feature and response distillation that handles different rain streak types with one pre-training weight. 
	
	\item 
	We design a simple and effective review module that converts the extracted residuals into old task rain streaks. The review module helps the derain network review the old task without viewing the old task data.
	
	\item 
	A large number of experiments demonstrate that the proposed method outperforms other SOTA methods in terms of rain removal effect and running speed, provided that only one model is used.
	
\end{itemize}

\section{Proposed Method}\label{sec:Method}

\subsection{The Overall Framework}
In this work, our goal is to remove different types of rain streaks by using one pre-training model. To this end, we propose a solution based on knowledge distillation. Fig.~\ref{fig:flow chart} illustrates the training flow of our proposed method. During the training phase, our approach consists of a student net (current network) $N^j$ and a teacher net (previous stage network) $N^{j-1}$. And during the evaluation phase, the student network $N^j$ completes the derain task alone. Their input is a video consisting of multiple consecutive frames. We divide the videos into batches, each batch consisting of 5 frames ($\{\X_{t-2},\X_{t-1},\X_{t},\X_{t+1},\X_{t+2}\}$), where $\X_{t}$ is the derain object and the other frames provide timing information for the model. The final output of the model is the rain-free background $\B$. The review network $\mathcal{R}_{e}$ generates a rain map of the old task,  whose input is the residuals extracted by the derain network and whose output is the rain map of the old task.
\subsection{Frame Grouping Module}

The temporal information is crucial for video deraining. This is because a real rainy video contains various types of rain streaks, while the background information remains constant. Leveraging this common background information can enhance the model's capability to extract rain streaks. Moreover, the running speed is a crucial factor to make the rain removal method feasible in practice. However, how to effectively and efficiently utilize the temporal information poses a challenging issue in video deraining.

To improve the model's ability for perceiving video information. We propose the Frame Grouping Module (FGM). 
Distinguishing from traditional U-Net architectures, we use two encoders to extract different temporal information depending on the frame rate. This can be expressed as:

\begin{equation}
	\begin{array}{l}
		\mathcal{F}_1=\Psi_1(\X_{t-1},\X_{t},\X_{t+1}),\\
		\mathcal{F}_2=\Psi_2(\X_{t-2},\X_{t},\X_{t+2}),\\
		\mathcal{F}=\mathcal{F}_1+\mathcal{F}_2,
	\end{array}
\end{equation}
where $\Psi_1$ and $\Psi_2$ denote the encoder and $\mathcal{F}$ represents the high-dimensional features extracted by the encoder. This allows our network to quickly discover the differences between the central frame and other frames, thus improving the high-dimensional feature quality.

\subsection{Iterative Feature and Response Distillation}\label{subsec:FKD}

Most existing derain methods require the use of different pre-trained models to achieve optimal rain removal when dealing with different rain streak types. To address the above issues, we propose an iterative feature and response distillation training scheme. Given a training sample of $\mathcal{D}_{all}=\{(\X^j, \Y^j)\}_{j=1}^K $, where $\{\X^1, \X^2, ..., \X^K\}$ represents the different types of rain video and $\{\Y^1, \Y^2, ..., \Y^K\}$ represents the corresponding clean background frame. $K$ denotes the total number of rain types. The training samples at stage $j$ can be denoted as $\mathcal{D}^j=\{( \X^j, \Y^j)\} $. Once the model can handle the rain streaks in the current dataset, we will use a new type of rain streak as the dataset $\mathcal{D}^{j+1}$ and start the next training stage.

When $j = 1$, we initialize our derain models by conventional training methods, and the loss function can be:
\begin{equation}
	\begin{array}{l}
		\theta_t^1=\{\X_{t-2}^1,\X_{t-1}^1,\X_{t}^1,\X_{t+1}^1,\X_{t+2}^1\},\\
		\mathcal{L}_{C}=\mathcal{L}(\mathcal{N}^1(\theta_t^1),\Y_{t}^1),
	\end{array}
\end{equation}
where $(\X_{t-2}^1, ..., \X_{t+2}^1, \Y_t^1)$ indicates the data in $\mathcal{D}^1$, $\theta_t^1$ represents the set of input frames, $\mathcal{N}^1$ denotes the derain model at stage $1$ and $\mathcal{L}$ is the loss function which consists of the $L_1$ norm and negative SSIM losses ($\mathcal{L}=\sigma_1 *\ell_{ssim}+\sigma_2 *\ell_1$, where $\ell_1$ denotes the $L_1$ norm, $\ell_{ssim}$ denotes the negative SSIM loss, and $\sigma_1$ and $\sigma_2$ denote the corresponding loss weights).

When $j > 1$, the model inherits the parameters and loss function from the previous stage and starts the next training phase.

To obtain a general video derain model, we utilize response knowledge distillation to constrain the optimization of the current model, i.e.,
\begin{equation}
	\begin{array}{l}
		\mathcal{L}_{RKD}=\mathcal{L}(\mathcal{N}^j(\theta^j_t),\mathcal{N}^{j-1}(\theta^j_t)).
	\end{array}
\end{equation}

This loss requires that the current model must approximate the old model's output and thus retain the old task data. To further constrain the current model optimization, we introduce feature knowledge distillation with the following loss:

\begin{equation}
	\begin{array}{l}
		\mathcal{L}_{FKD}=\mathcal{L}(\mathcal{\F}^{j},\mathcal{\F}^{j-1}),
	\end{array}
\end{equation}
where $\mathcal{F}^{j}$ denotes the features extracted by the $j$-stage model encoder. The loss requires the current model's encoder to extract features as close as possible to those extracted by the old task model. 

\begin{figure}[t]
	\centering 
	\begin{tabular}{c}
		\includegraphics[width=0.47\textwidth]{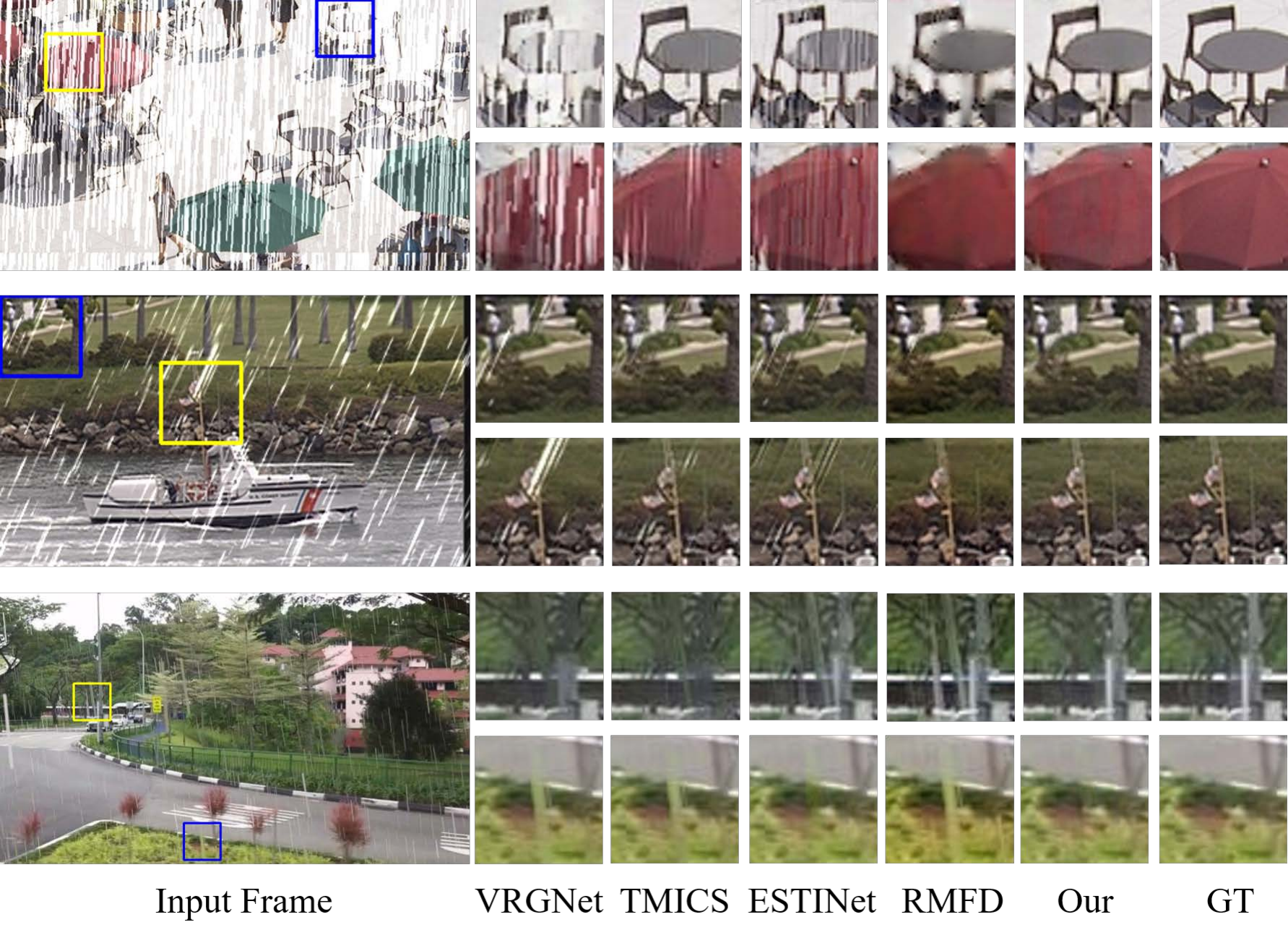}
	\end{tabular}
	\vspace{-3mm}
	\caption{Comparing with other methods on RainSynComplex25, RainSynLight25 and NTU datasets. }
	\label{fig:vision_complex} 
\end{figure}

\begin{figure}[t]
	\centering 
	\begin{center}
		\begin{tabular}{c@{\extracolsep{0.2em}}c@{\extracolsep{0.2em}}c}
			\includegraphics[width=0.15\textwidth]{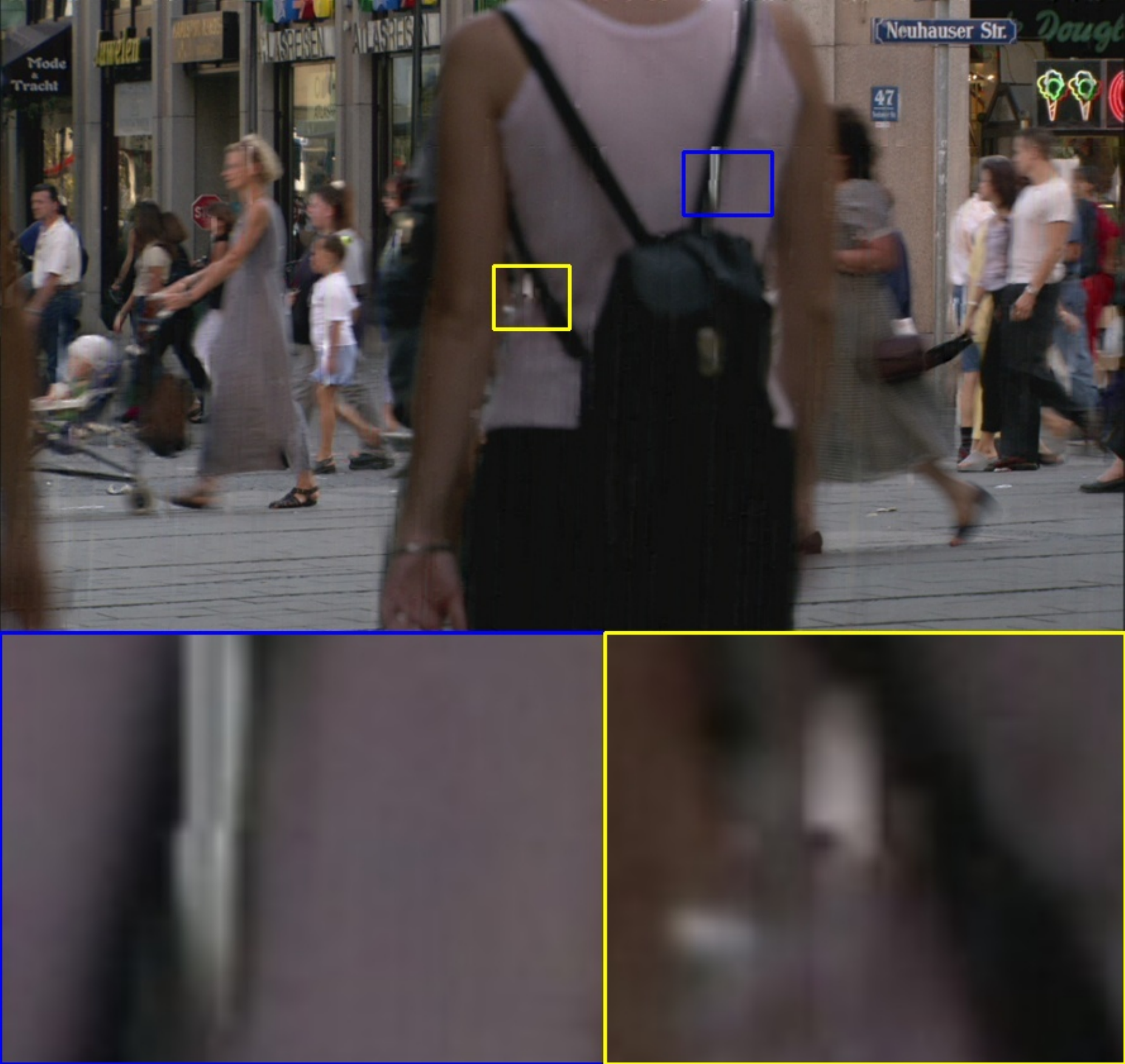}&
			\includegraphics[width=0.15\textwidth]{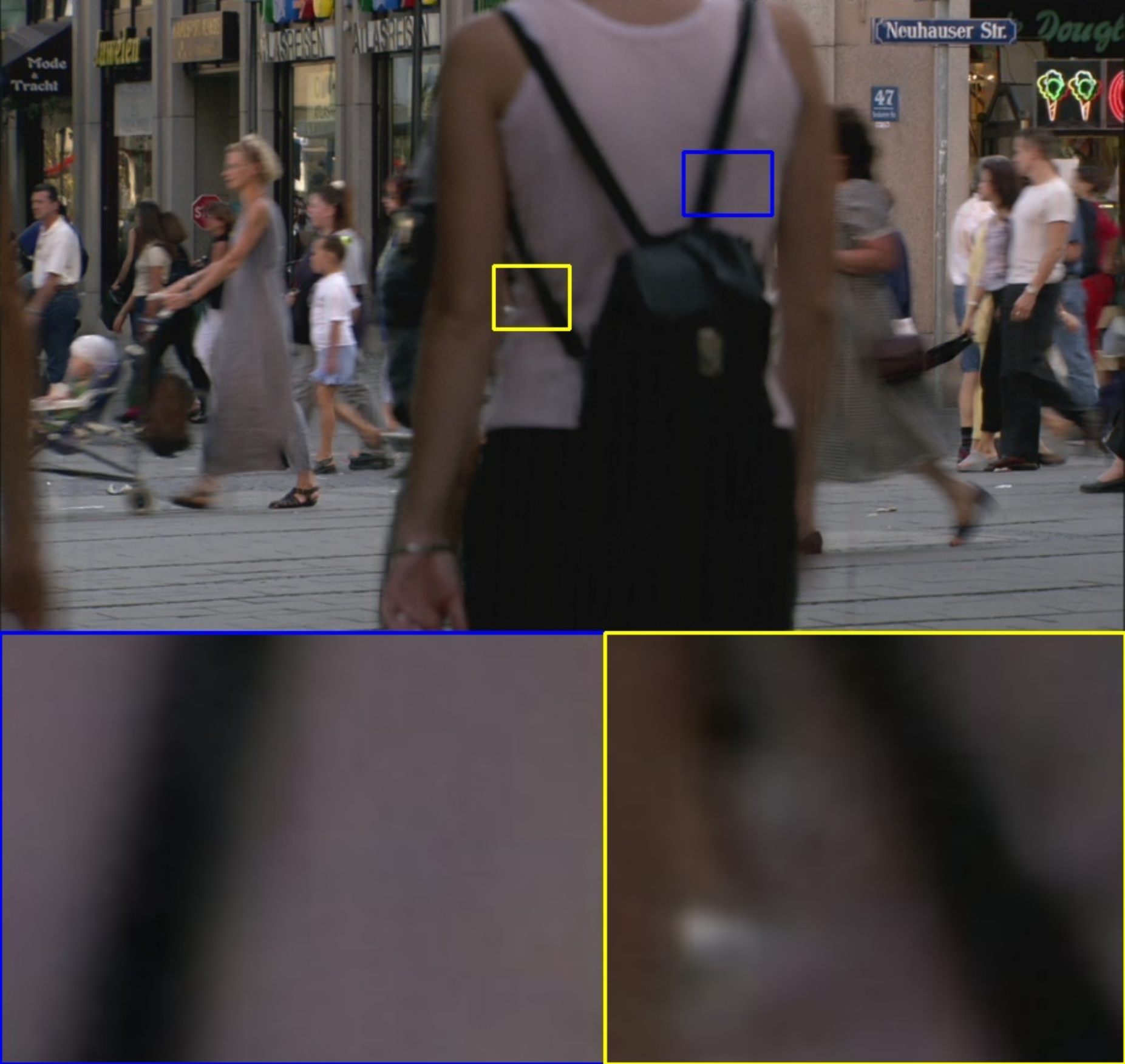}&
			\includegraphics[width=0.15\textwidth]{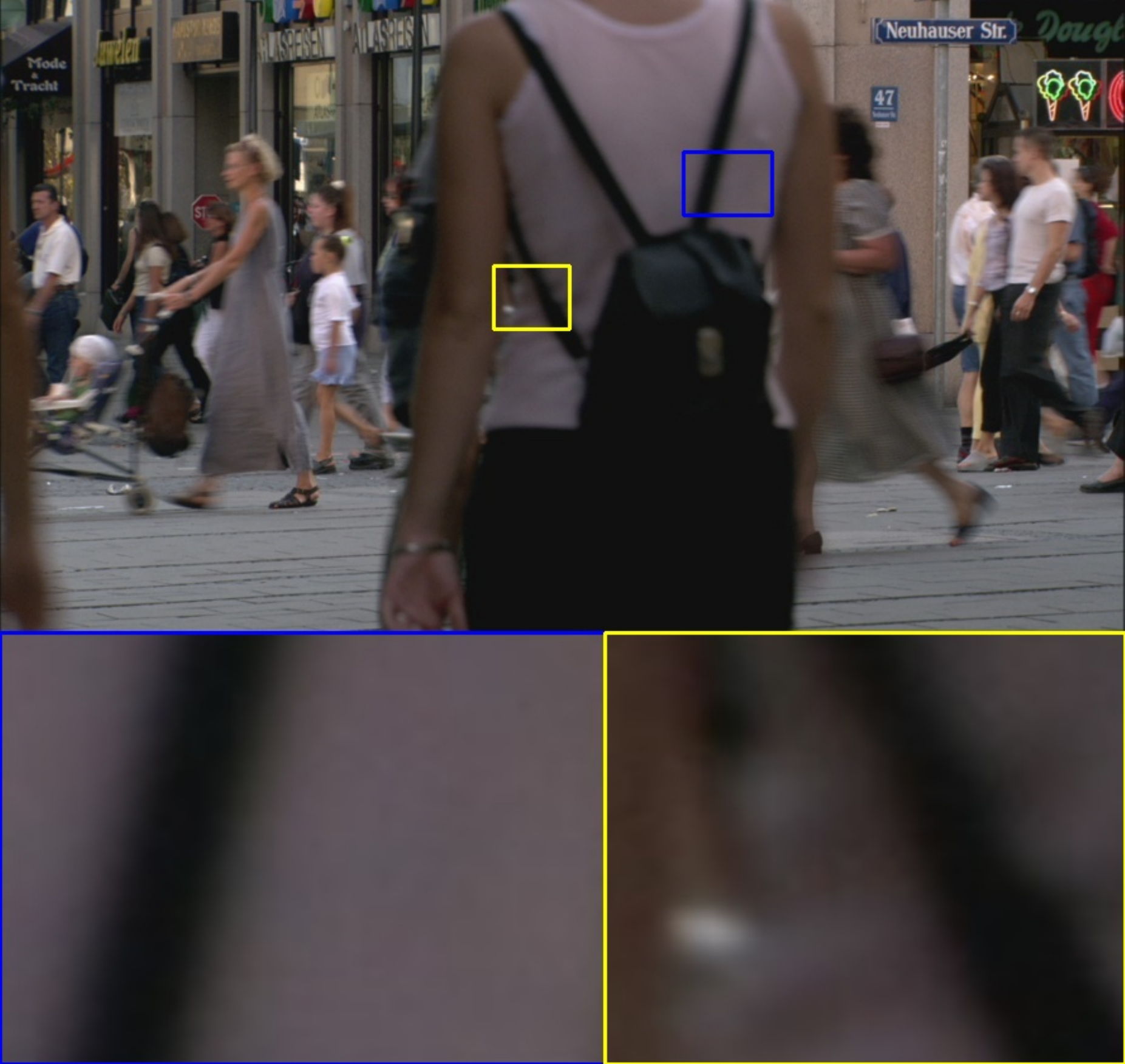}\\
			\footnotesize (a) w/o FGM & \footnotesize (b) w FGM &\footnotesize GT 
		\end{tabular}
	\end{center}
	\vspace{-3mm}
	\caption{We validate in our ablation experiments that grouping video frames together for processing helps the model extract rain streak information.}\label{fig:FG} 
\end{figure}

\begin{figure}[t]
	\centering 
	\begin{center}
		\begin{tabular}{c}
			\includegraphics[width=0.46\textwidth]{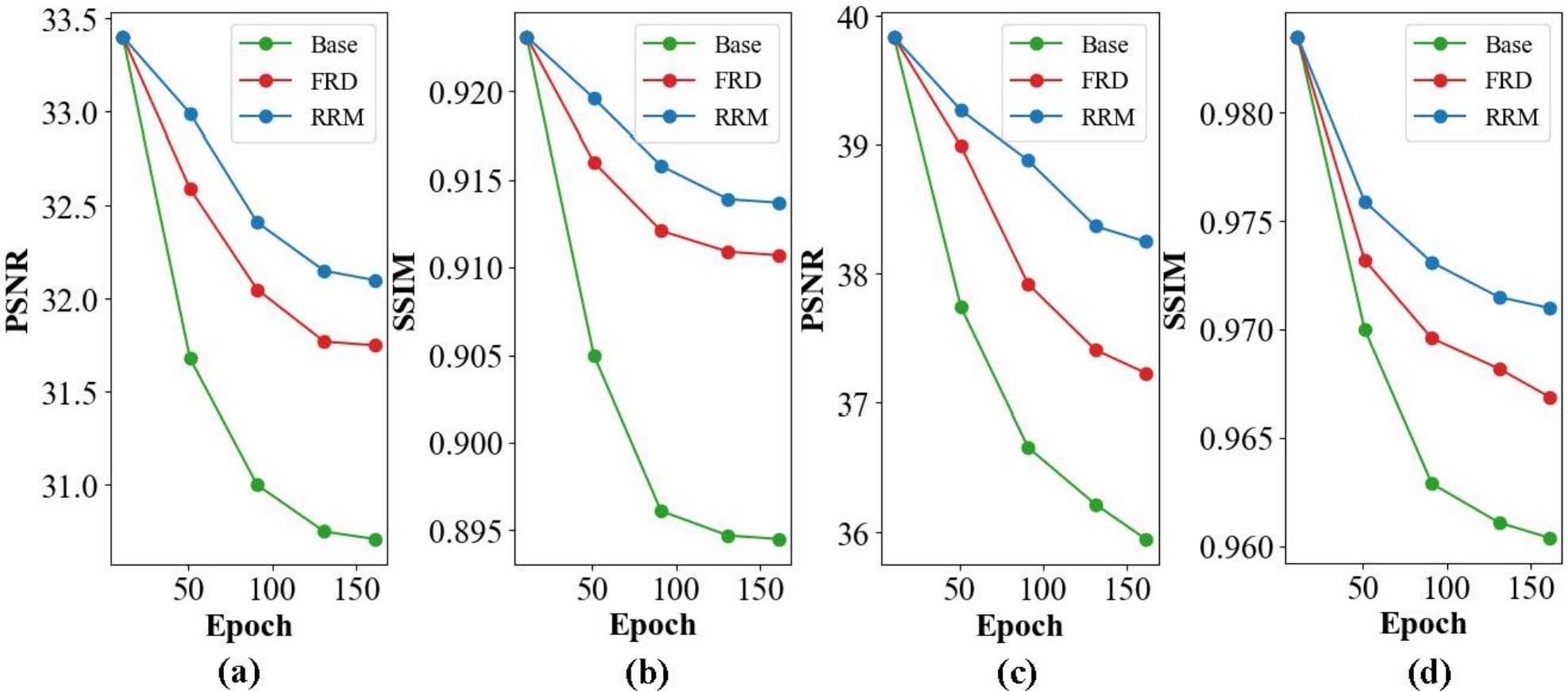}
		\end{tabular}
	\end{center}
	\vspace{-3mm}
	\caption{The degradation of the model's performance on the old tasks (Complex and Light dataset) as the model starts to learn the new dataset (NTU Dataset).  The decay of PSNR and SSIM on RainSynComplex25 are presented in (a) and (b), respectively. The results for RainSynLight25 are also presented in (c) and (d), respectively.}\label{fig:down} 
	
\end{figure}
\subsection{Rain Review Module}\label{subsec:RRM}

To avoid catastrophic forgetting problems, we design a simple and effective rain streak generation module called Rain Review Module (RRM). This module uses the background residuals extracted by the previous model to generate rain streaks of the old task. The module improves the performance of the model on both new and old tasks by data augmentation. To avoid providing unlearned rain knowledge, the review network is trained using the same approach as the derain network. 
We use the $j$-stage training process ($j>1$) as an example to show how the review network assists the derain network in recalling old knowledge. The review network takes the residuals extracted from the previous stage network as input and outputs a rain map with information about the old task rain streaks:
\begin{equation}
	\begin{array}{l}
		\S=\mathcal{R}_e(\R^{j-1}_t),
	\end{array}
\end{equation}
where $\mathcal{R}_e$ denotes the review network, $\R^{j-1}_t$ is the residuals extracted by the previous stage network, and $\S$ means the rain streak map generated by the rain review module. Next, we fuse the rain streak map with the background extracted from the previous stage network to obtain a completely new rain map. To enhance the review effect, we use affine variation to augment the data for the rain streak map.
\begin{equation}
	\begin{array}{l}
		\tilde{\X}=\mathcal{A}(\S) +\B^{j-1}_t,
	\end{array}
\end{equation}
where $\mathcal{A}$ denotes the affine variation, $\B^{j-1}_t$ indicates the background map output from the old task network and $\tilde{\X}$ denotes the newly synthesized rain map. After that, we let the current network remove the rain streak in $\tilde{\X}$:
\begin{equation}
	\begin{array}{l}
		\tilde{\B}_r=\mathcal{N}^j(\X_{t-2},\X_{t-1},\tilde{\X},\X_{t+1},\X_{t+2}),\\
		\mathcal{L}_{R}=\mathcal{L}(\tilde{\B}_r,\Y^{j}_t),
	\end{array}
\end{equation}
where $\mathcal{N}^j$ denotes the current stage network, $\tilde{\B}_r$ represents the background of the current network recovered according to the new rain map, and $\Y^{j}_t$ indicates the corresponding ground truth. Through the constraint of loss, $\mathcal{L}_{R}$, the network completes the recall of the old task data. In addition, by training the old task data together with the new task data, the model further learns common features in the different rain streaks. This helps the model to learn new rain streaks knowledge as well.

\subsection{Overall Loss}\label{subsec:loss}
We show all loss functions in Section~\ref{subsec:FKD} and Section~\ref{subsec:RRM}. When training the derain network, the total losses are:
\begin{equation}
	\begin{array}{l}
		\mathcal{L}_{\mathcal{N}}=\lambda_1\mathcal{L}_{C}+\lambda_2\mathcal{L}_{RKD}+\lambda_3\mathcal{L}_{FKD}+\lambda_4\mathcal{L}_{R},\\
	\end{array}
\end{equation}
where $\lambda_1$ to $\lambda_4$ denotes loss weights. 
The review network training loss is similar to the derain network, except that it does not have a review module:
\begin{equation}
	\begin{array}{l}
		\mathcal{L}_{Re}=\lambda_1\mathcal{L}(\mathcal{\F}_{r}^{j},\mathcal{\F}_{r}^{j-1})+\lambda_2\mathcal{L}(\S^{j},\S^{j-1})+\\
		\lambda_3\mathcal{L}(\S^{j},\mathcal{G}(\X^{j}_{t}-\Y^{j}_{t})),
	\end{array}
\end{equation}
where $\mathcal{F}_{r}^{j}$ denotes the features extracted by the $j$-stage review network and $\S^{j}$ denotes the rain map it produces, $\Y^{j}_{t}$ indicates the corresponding ground truth and $\mathcal{G}$ denotes the graying process.

\begin{table}[t]
		\centering
		\caption{Ablation study on different settings with averaged PSNR/SSIM. }
		\label{table:Ablation}
		\setlength{\tabcolsep}{1.5pt}
		\begin{tabular}{cccc}
			\hline
			\toprule
			Settings & (a) & (b) & (c) \\
			\midrule
			Base  & $\checkmark$ & $\checkmark$ & $\checkmark$  \\
			FRD  &              & $\checkmark$ & $\checkmark$  \\
			RRM  &            &              & $\checkmark$  \\
			\midrule
			{SynComplex25}  & 30.71/0.8945 & 31.75/0.9107 & \bf{32.10/0.9137}   \\
			{SynL25 }   & 35.95/0.9604 & 37.23/0.9669 & \bf{38.25/0.9710}  \\
			\bottomrule
			
		\end{tabular}
\end{table}

\begin{table}[t]
	\begin{center}
		\centering
		\caption{Ablative analysis of Frame Grouping Modules}
		\label{table:FG}
			\begin{tabular}{ccc}
				\hline
				\toprule
				\multirow{2}{*}{Settings} & w/o FGM  & w FGM  \\
				\cline{2-3}
				& PSNR / SSIM & PSNR / SSIM \\
				\midrule
				SynComplex25 & 32.01 / 0.9124 & 32.10/0.9137 \\
				SynL25  & 37.95 / 0.9711 & 38.25/0.9710  \\
				\bottomrule
			\end{tabular}
	\end{center}
\end{table}

\begin{table*}[htb!]
	\begin{center}
		\centering
		\caption{Averaged PSNR and SSIM results among different rain streaks removal methods on four different video datasets, i.e., RainSynLight25 (short for D1), RainSynComplex25 (D2) and  NTURain (D3). }
		\label{table:comparison sota} 
		\setlength{\tabcolsep}{1.3pt}
		\begin{tabular}{cc|cccccccccccc}
			\hline
			\toprule
			\multicolumn{2}{c|}{Datasets} & FastDeRain & JORDER & J4R-Net & SpacCNN & MPRNet & VRGNet & DualFlow & TMICS&  RMFD  & ESTINet & MFDNN & Ours\\ 
			\hline
			\multirow{2}{*}{D1} & PSNR   & 29.42& 30.37   & 32.96 & 32.78 & 33.47 & 34.53 & 35.80 &36.65 & 36.99 & 36.12 &37.47 &  {\textbf{38.79}} \\
			& SSIM & 0.8683 & 0.9235  & 0.9434 & 0.9239 & 0.9683 & 0.9630 & 0.9622 &0.9689  & 0.9760  & 0.9631 & 0.9550 & {\textbf{0.9734}} \\
			\hline 
			
			\multirow{2}{*}{D2} & PSNR  & 19.25 & 20.20 & 24.13 & 21.21 & 27.92 & 28.69 &  27.72 &29.49 & 32.70 & 28.48 &31.95 &  {\textbf{32.78}}\\
			
			& SSIM  & 0.5385 & 0.6335  & 0.7163  & 0.5854 & 0.9112  & 0.8832 &  0.8239 & 0.8933 & \bf{0.9357}  & 0.8242 &0.7966 & 0.9171 \\
			\hline
			
			\multirow{2}{*}{D3} & PSNR  & 30.32 & 32.61& 32.14 & 33.11 & 35.60 & 35.20 & 36.05 &37.38 & 38.92  & 37.48 & 36.27 & \textbf{39.55}\\
			
			& SSIM & 0.9262& 0.9482 & 0.9480 &0.9475 & 0.9752 & 0.9767 & 0.9676 &0.9704 & 0.9764  & 0.9700 & 0.9657 & \textbf{0.9824} \\
			
			\bottomrule
		\end{tabular}
	\end{center}
\end{table*}

\begin{figure*}[t]
	\centering 
	\begin{center}
		\begin{tabular}{c@{\extracolsep{0.2em}}c@{\extracolsep{0.2em}}c@{\extracolsep{0.2em}}c@{\extracolsep{0.2em}}c@{\extracolsep{0.2em}}c}
			
			\includegraphics[width=0.155\textwidth]{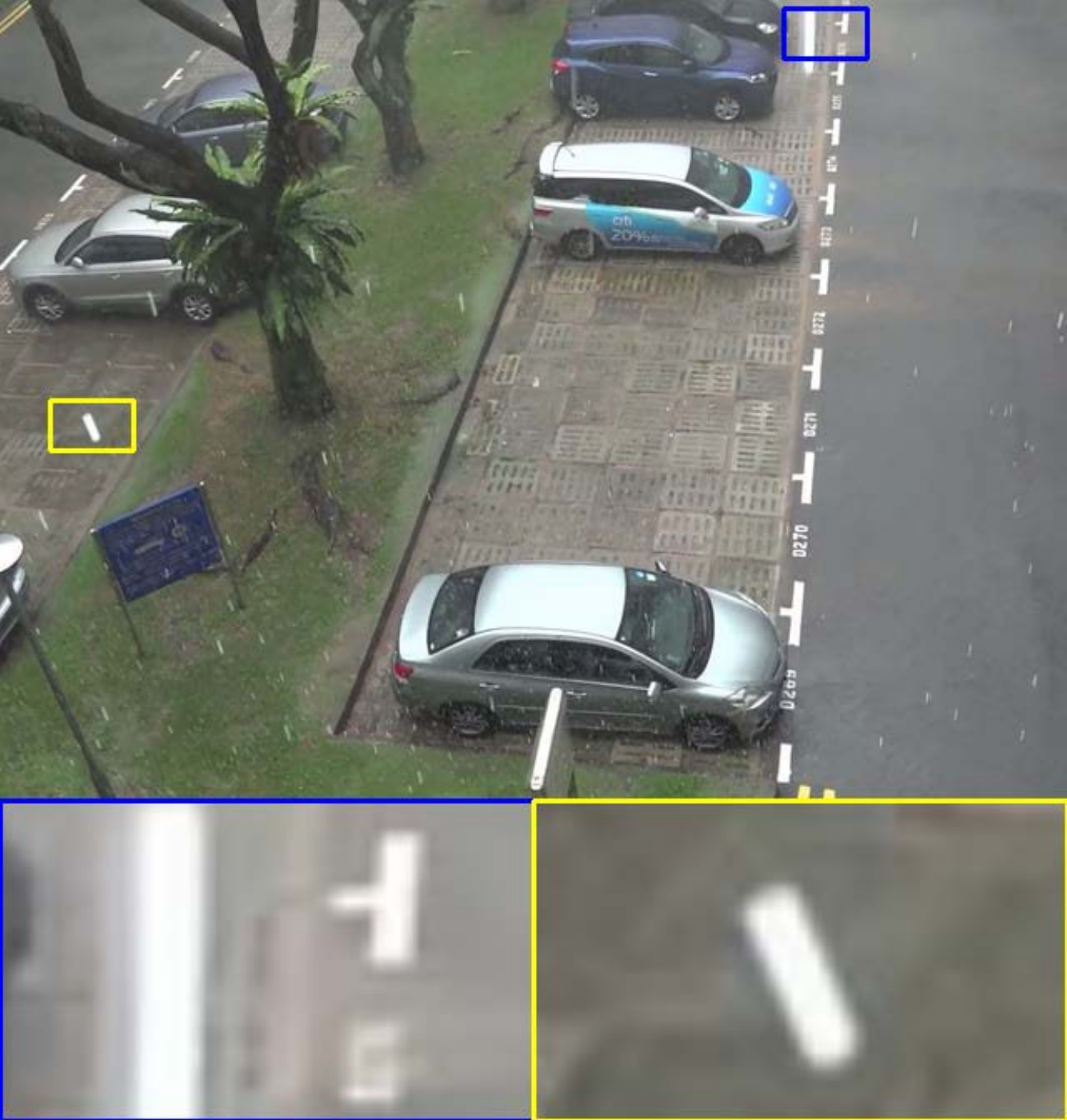}&
			\includegraphics[width=0.155\textwidth]{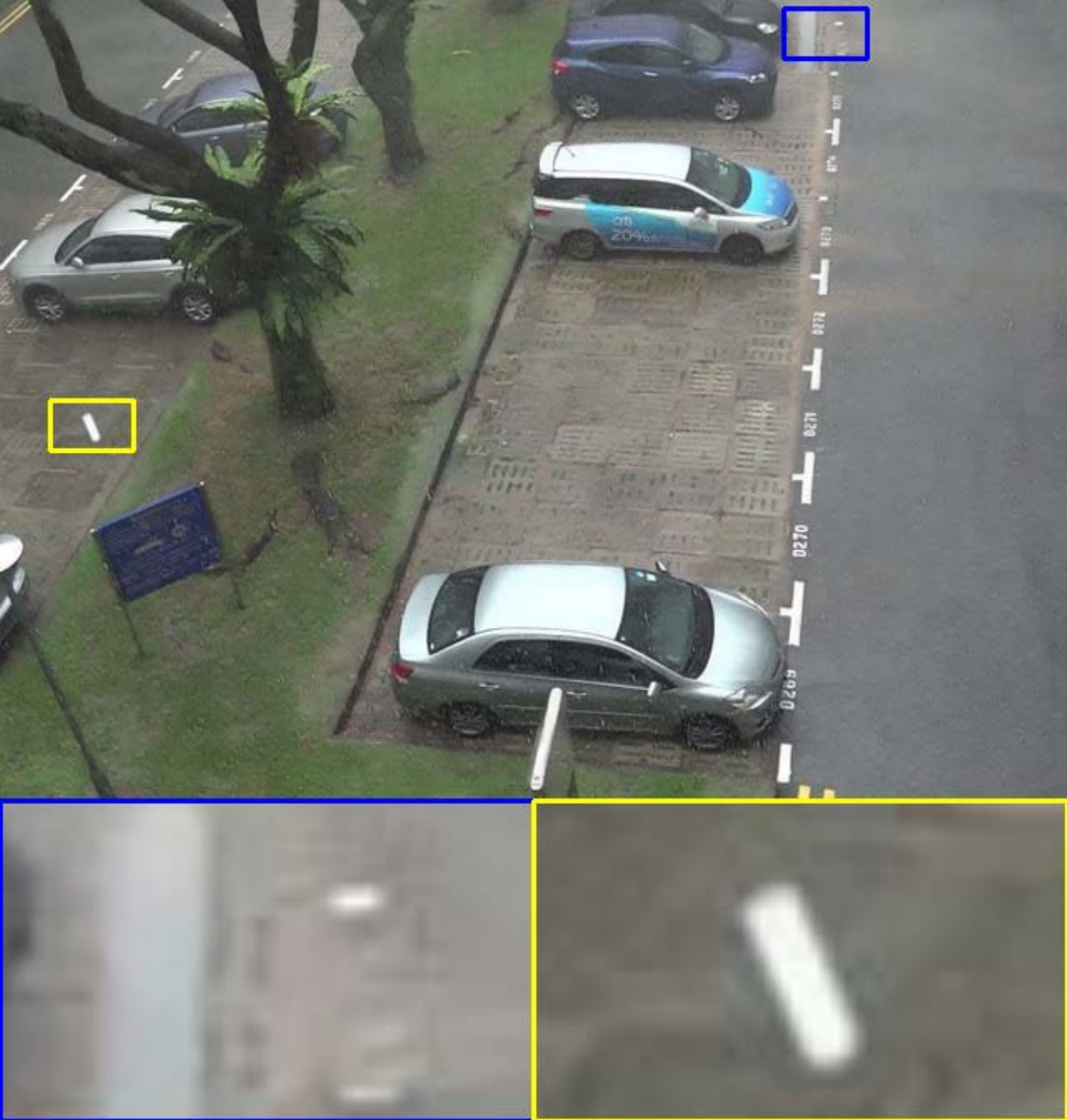}&
			\includegraphics[width=0.155\textwidth]{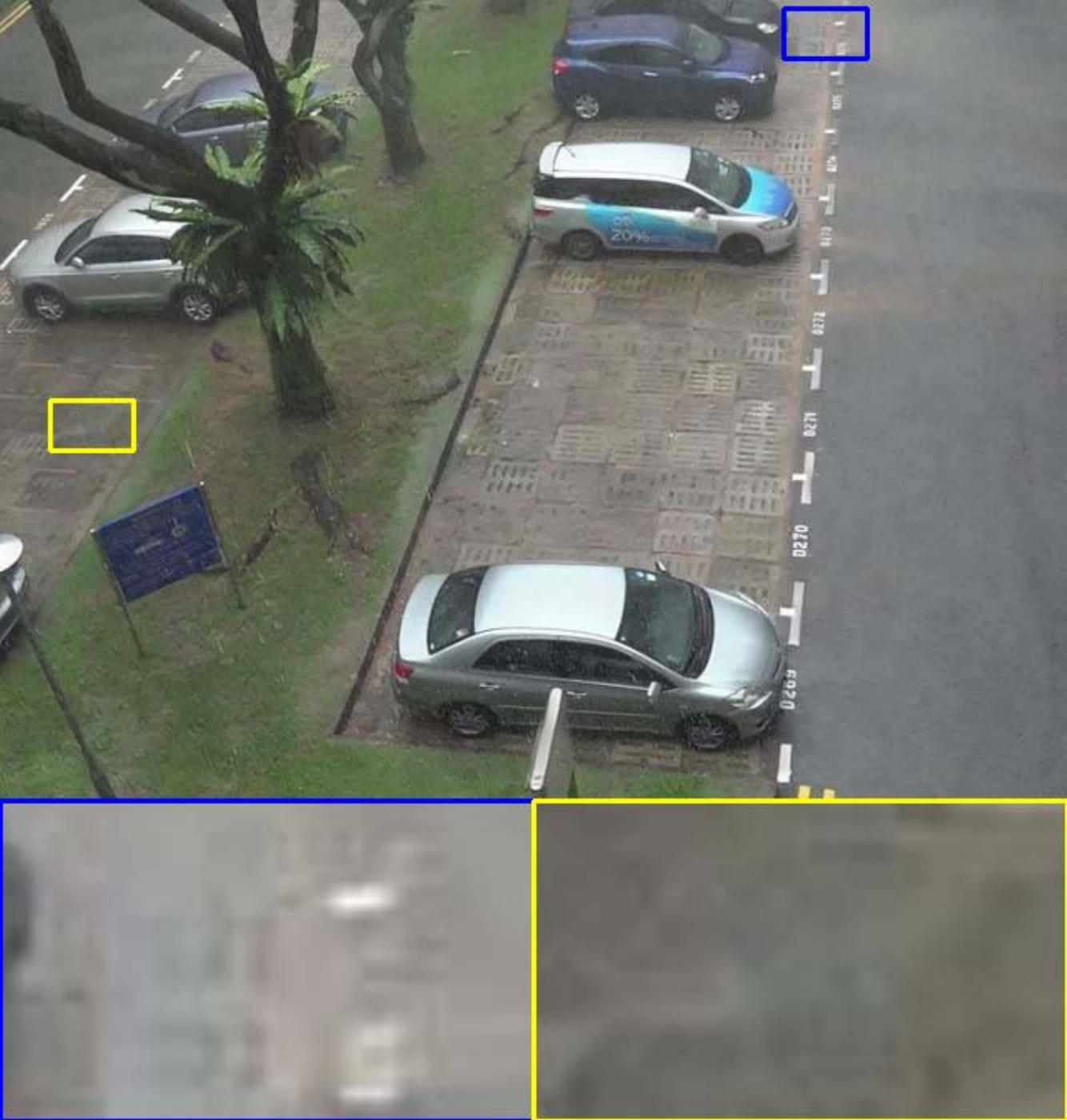}&
			\includegraphics[width=0.155\textwidth]{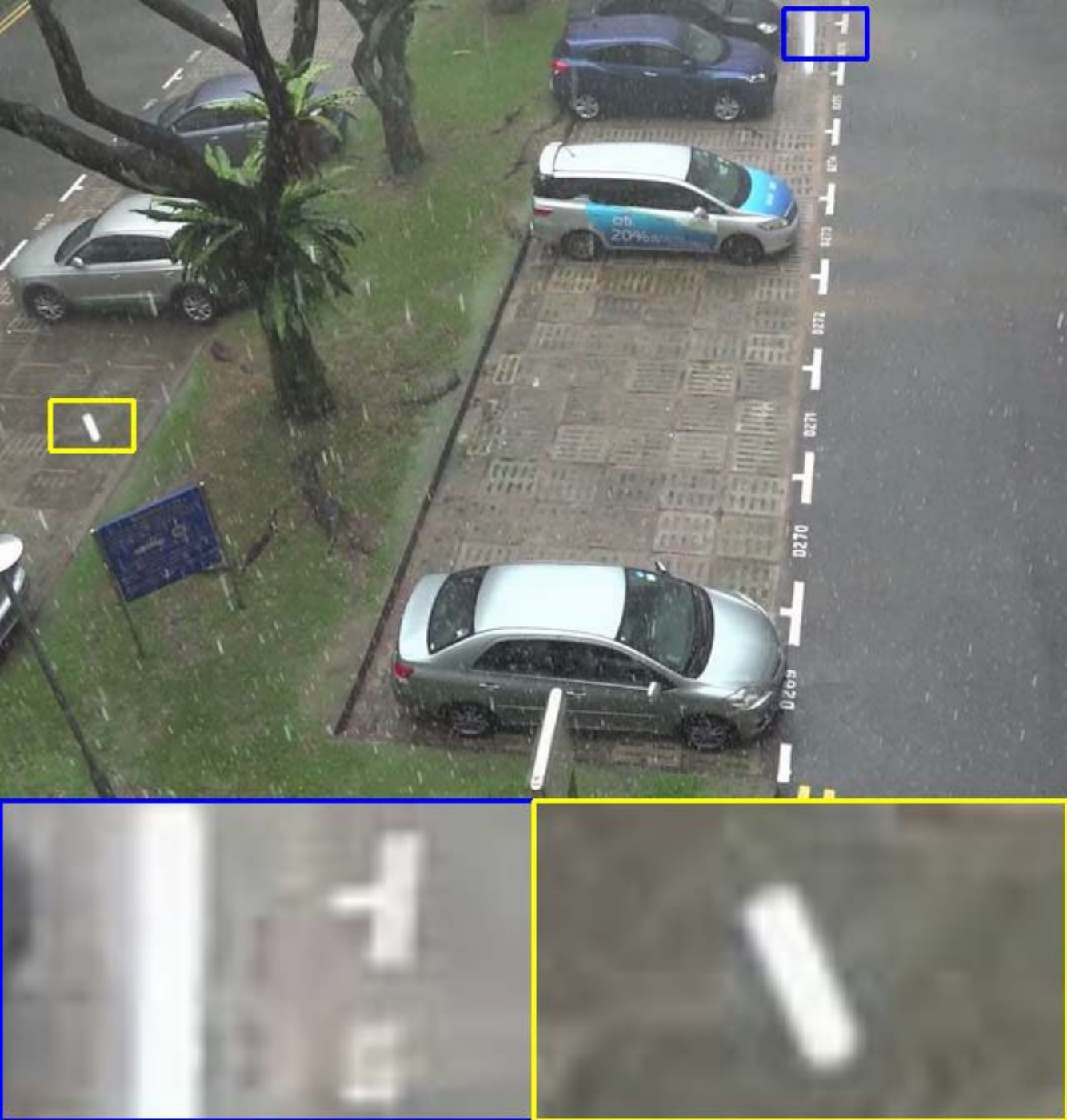}&
			\includegraphics[width=0.155\textwidth]{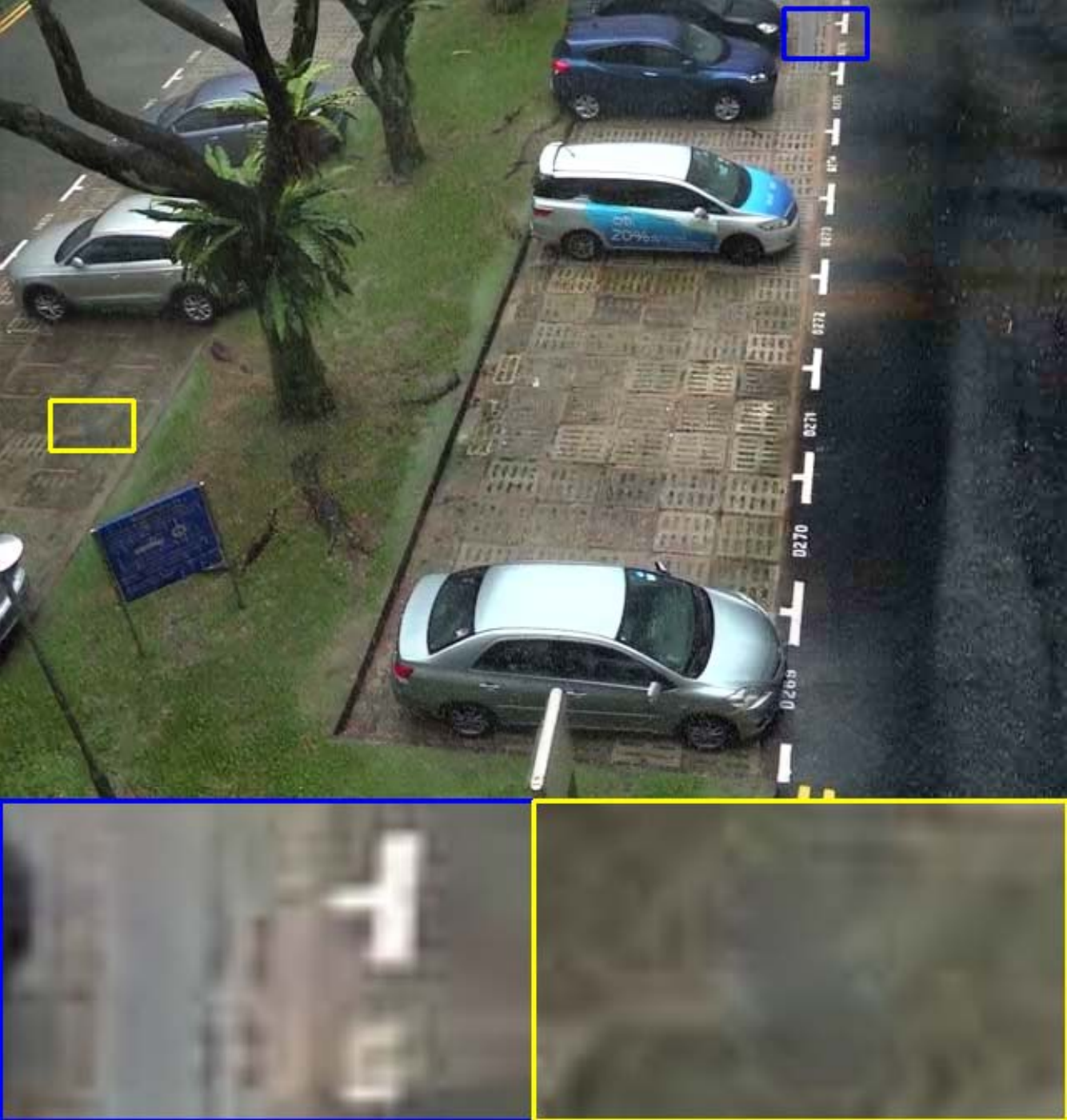}&
			\includegraphics[width=0.155\textwidth]{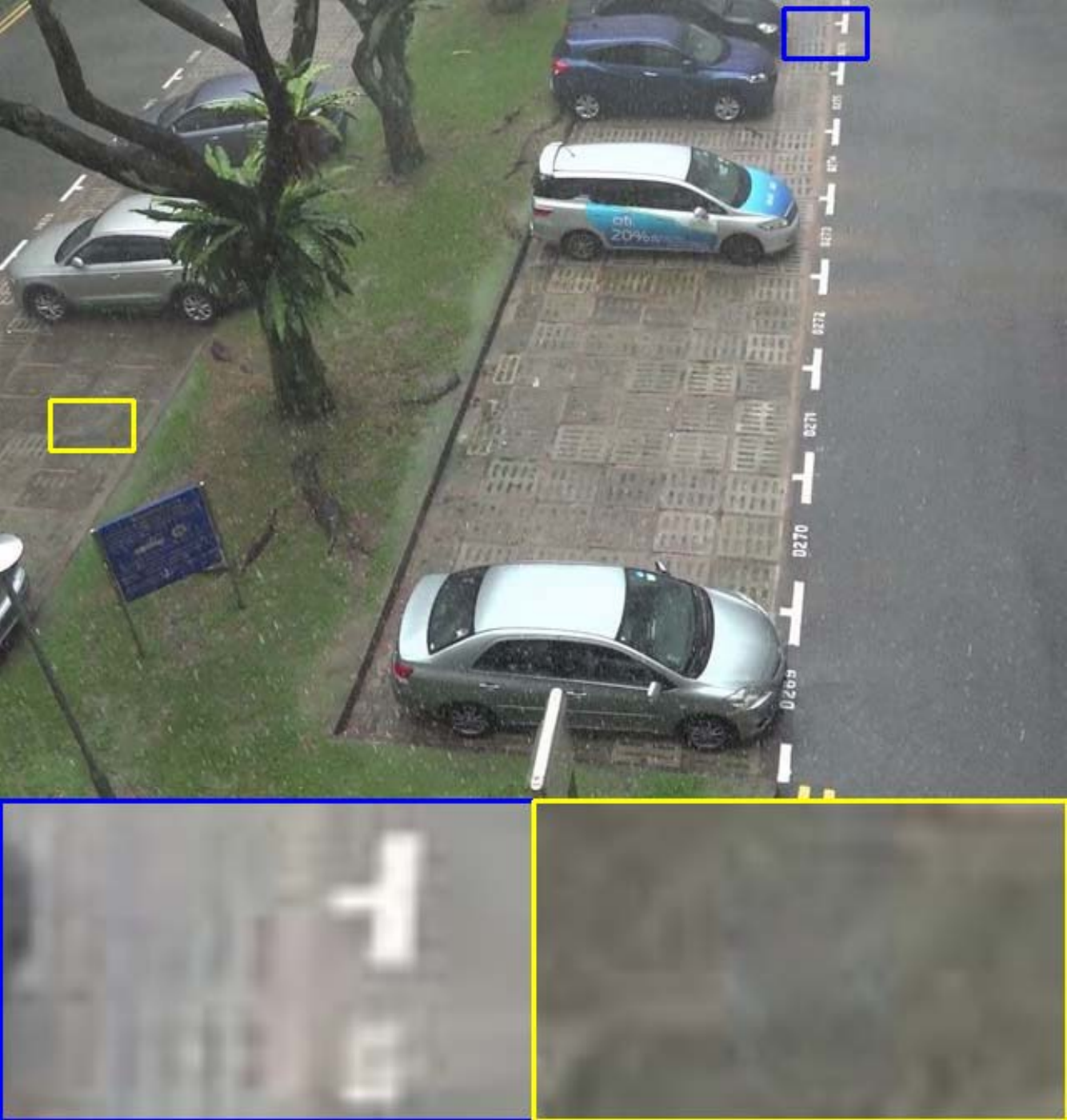}\\
			
			\includegraphics[width=0.155\textwidth]{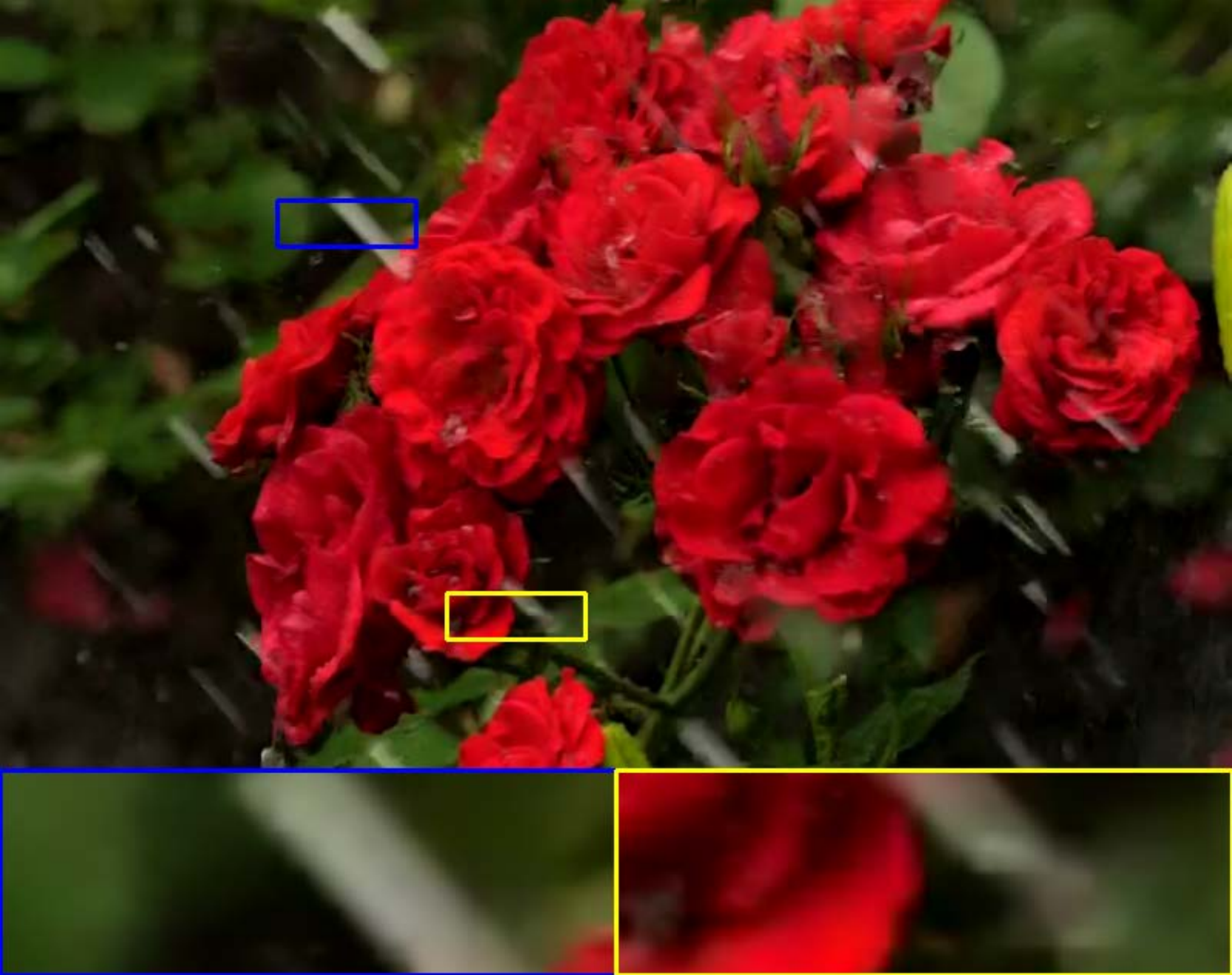}&
			\includegraphics[width=0.155\textwidth]{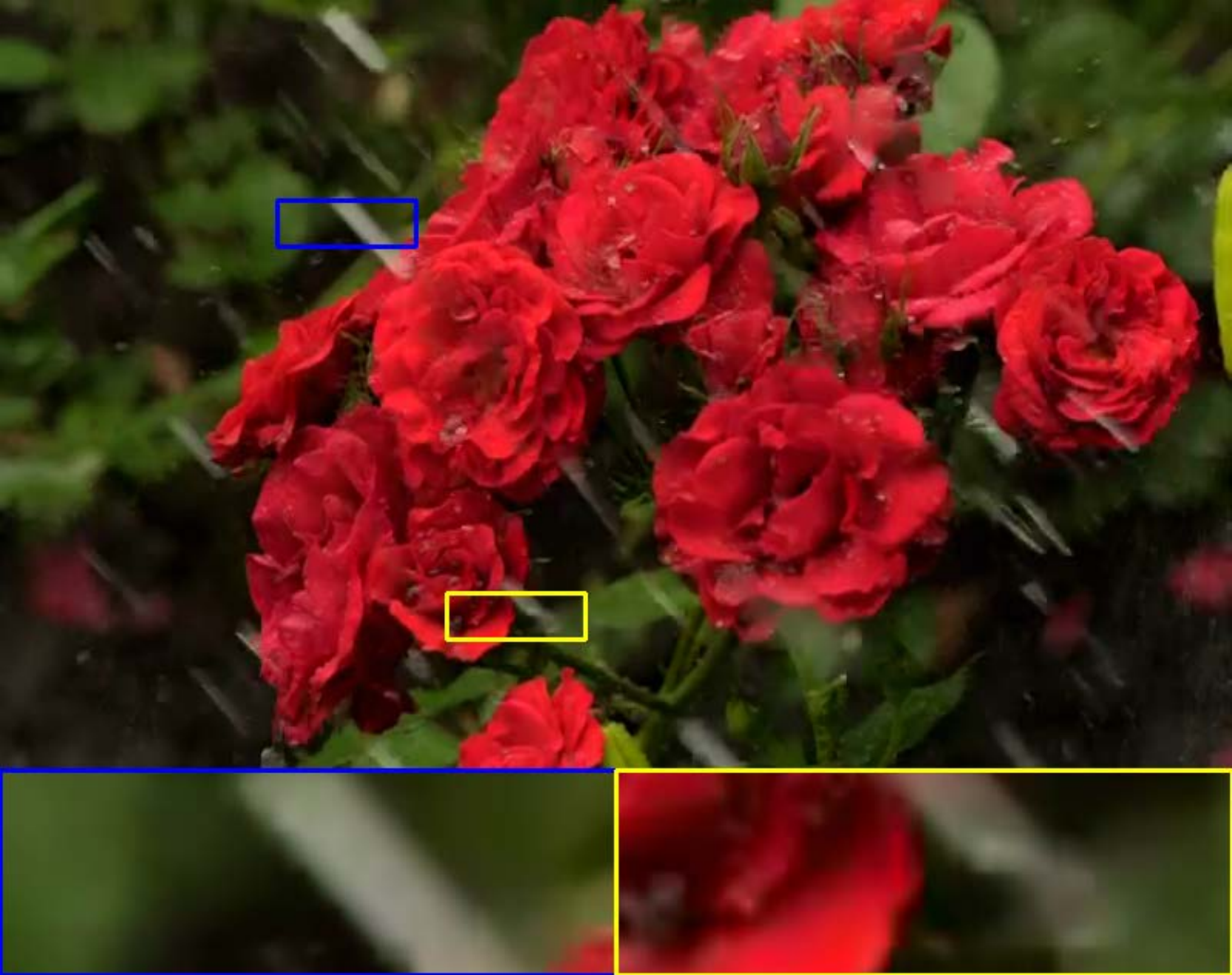}&
			\includegraphics[width=0.155\textwidth]{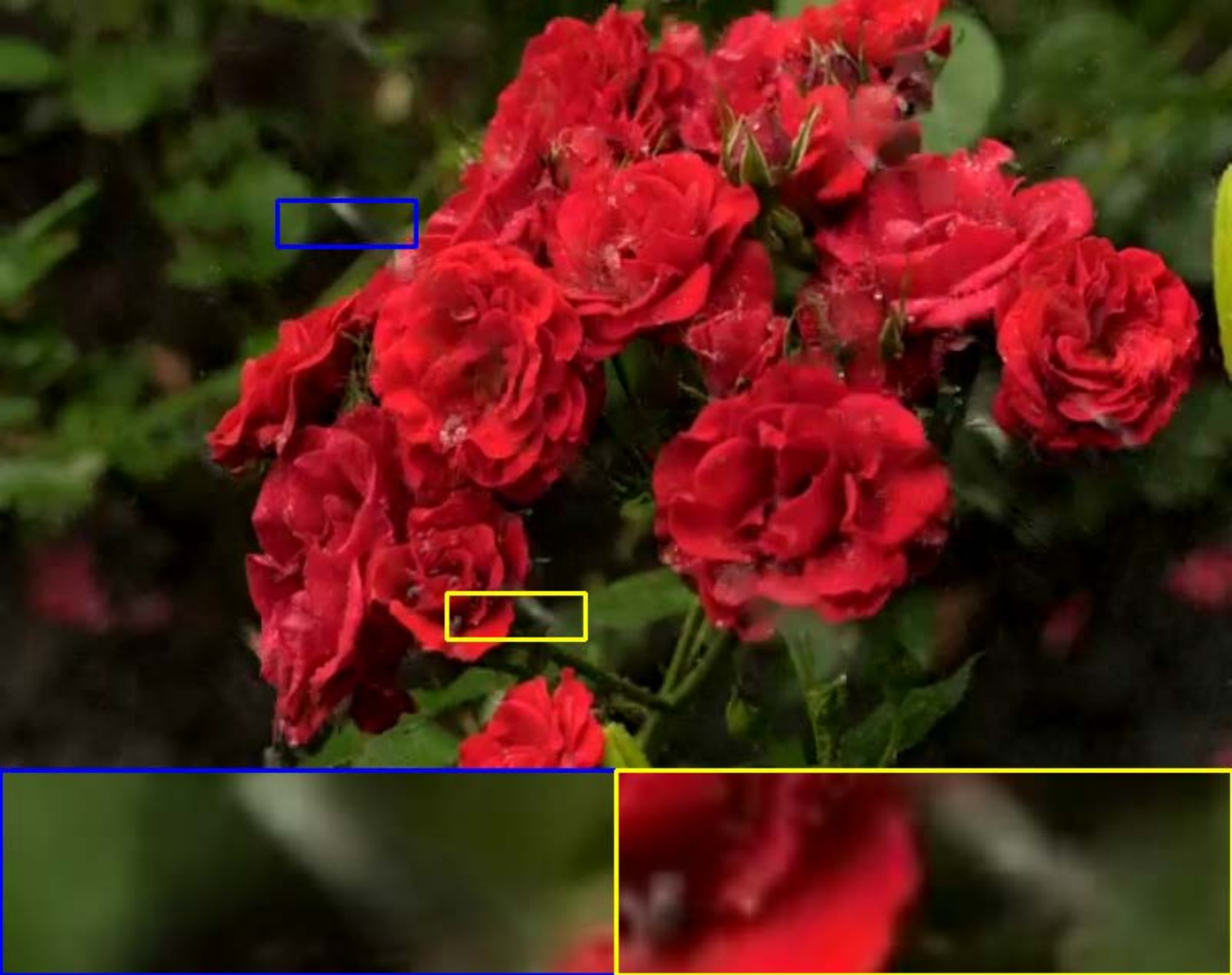}&
			\includegraphics[width=0.155\textwidth]{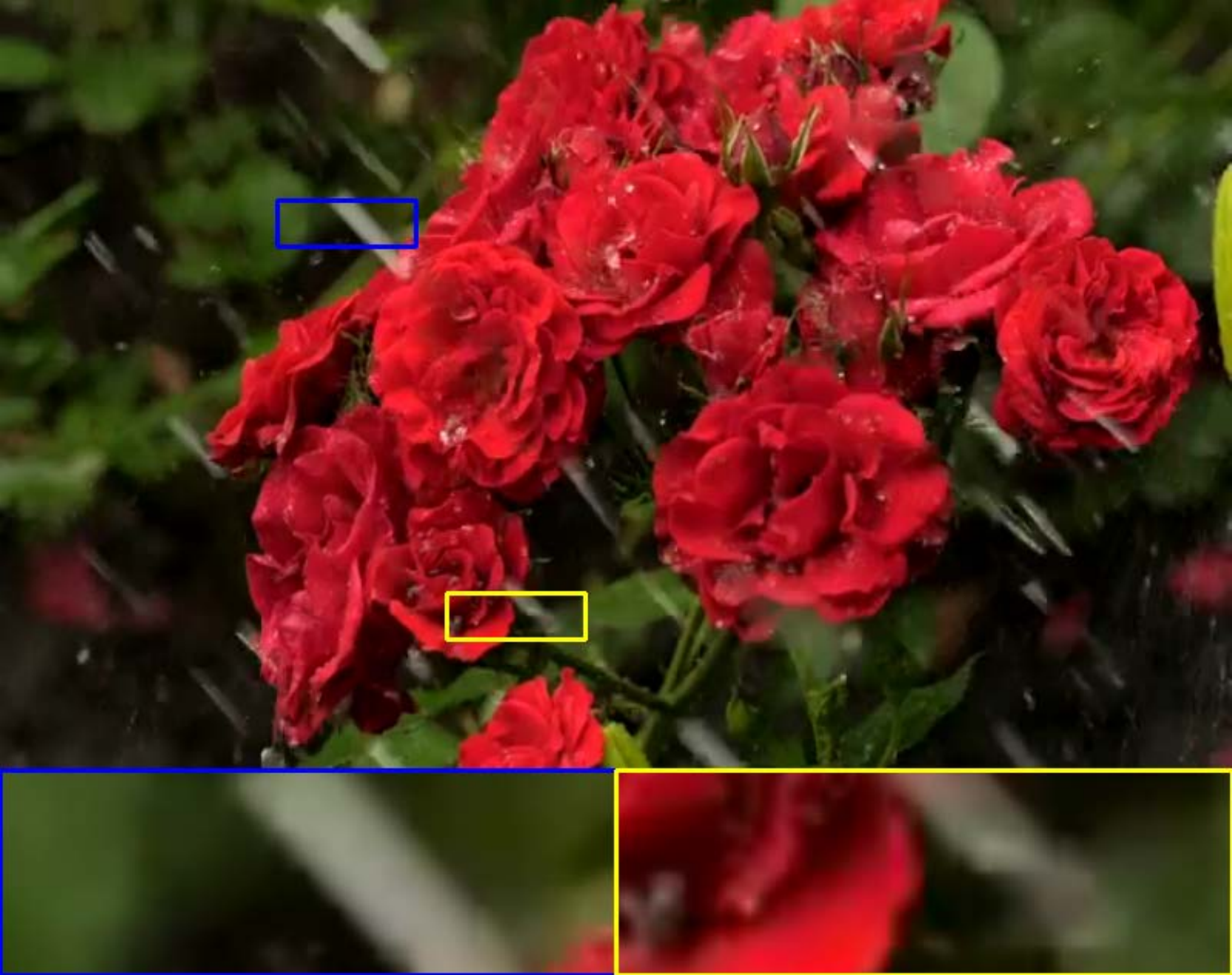}&
			\includegraphics[width=0.155\textwidth]{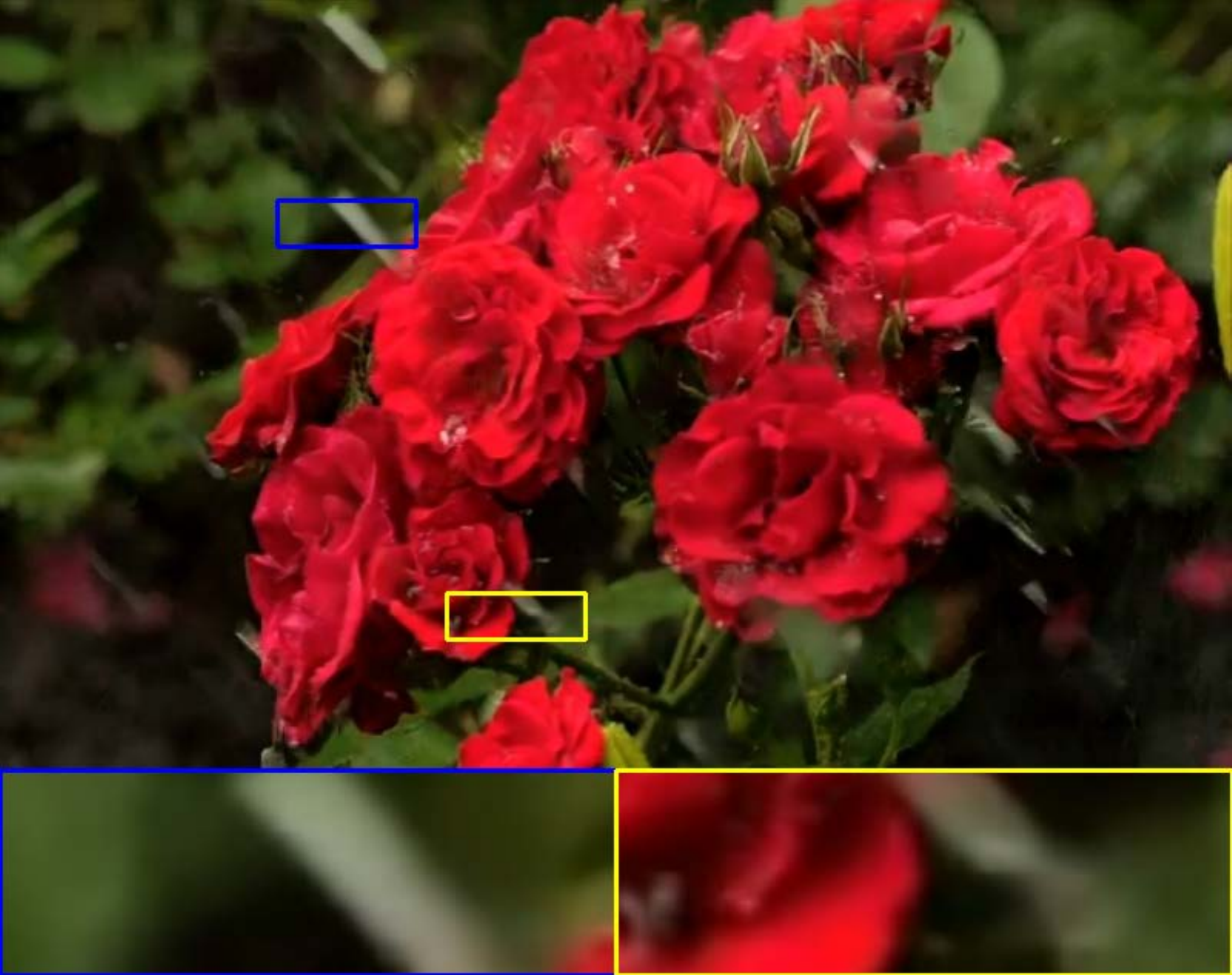}&
			\includegraphics[width=0.155\textwidth]{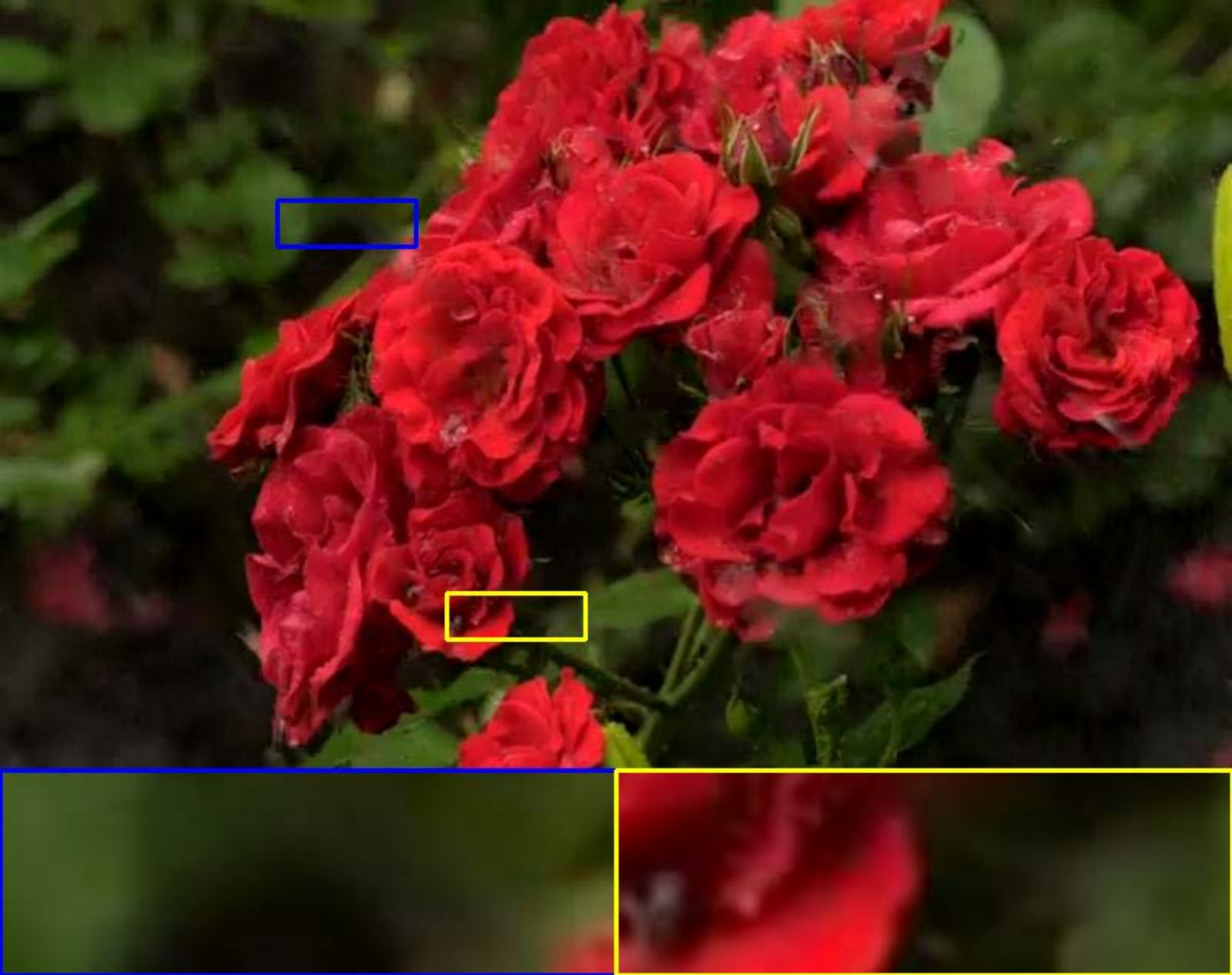}\\
			
			\footnotesize MPRNet &\footnotesize VRGNet  &\footnotesize TMICS   & \footnotesize ESTINet &\footnotesize RMFD & \footnotesize Ours 
		\end{tabular}
	\end{center}
	\vspace{-3mm}
	\caption{ comparing with other methods on real rainy video. 
	}\label{fig:vision_real} 
\end{figure*}

\section{Experimental Results}
\subsection{Implementation Details}
We compare the proposed method with the state-of-the-art method on three datasets, RainSynLight25~\cite{liu2018erase}, RainSynComplex25~\cite{liu2018erase}, NTU-Rain~\cite{chen2018robust}. 
We use the most widely evaluated metrics peak signal-to-noise ratio (PSNR) and structural similarity (SSIM) as quantitative evaluation metrics for all methods. The evaluation results are performed on the luminance channel. 

Each network is trained for 160 epochs with a learning rate of 0.001. The Adam optimizer is used, and the batch size is 1. We randomly clip all inputs to 240 $\times$ 240 size. The loss weights $\sigma_1$ and $\sigma_2$ are 1.1 and 0.75, respectively. And $\lambda_1$ to $\lambda_4$ are 0.5, 0.5, 1 and 1 respectively.

\begin{table}[t]
	\begin{center}
		\centering
		\caption{Comparisons on model complexity and running time. We use the same calculation method as ~\cite{yang2021recurrent} for video frames with a resolution of $832\times 512$.}
		\label{table:runtime} 
		\begin{tabular}{cccc}
			\hline
			\toprule
			Methos  & JORDER  & TMICS & DualFlow\\
			\midrule
			{Speed}    &  0.3346  & 0.6240  & 0.7627\\
			{Param}      &  4,169,024   & 8,215,160 & 4,466,694 \\		
			\hline
			\hline
			Methos    & RMFD & ESTINet  &  Ours \\ 
			Speed      & 0.4374 & 0.3104 & \textbf{0.103}\\
			Param     & 29,472,018 & 6,897,386  & 6,965,103  \\
			\bottomrule
		\end{tabular}
	\end{center}
\end{table}
\subsection{Ablation Study}

\textbf{Effectiveness of knowledge distillation.} To confirm whether feature and response distillation maintain the performance of the derain model on the old task, we compare the difference in performance on old tasks between the model with the addition of knowledge distillation and the base network. The results are shown in Table~\ref{table:Ablation}. The base network in the table indicates the derain module trained by conventional means. And FRD denotes a derain network trained by feature and response distillation. To ensure fairness, the same training setup is  used for each method and the experimental results are compared for the final epoch. It can be seen that the network with knowledge distillation still performs better on the old task compared to the base network.

\textbf{Effectiveness of rain review modules.} To verify the effectiveness of the rain review module, we compare the impact of with and without the review module on the derain performance. The results are presented in Table~\ref{table:Ablation}. It can be seen that the review module strongly improves the rain removal performance of the model.

To further illustrate the effect of the feature and response distillation and the rain review modules on the old task, we show the performance degradation of the model on the old task (RainSynLight and RainSynComplex) for different settings in Fig.~\ref{fig:down}. It can be seen that the base network quickly forgets how to handle the old task, while the addition of data distillation effectively maintains the performance of the model on the original task. And with the addition of the review module, the model reaches convergence with only a small cost, which demonstrates the effectiveness of our proposed module.

\textbf{Effectiveness of frame grouping modules.} To verify whether the model needs to group the inputs according to frames, we compare the results with and without grouping, and the experimental results are shown in Table~\ref{table:FG}. The effect of grouping is 0.3 dB higher than that of direct input on the RainSynLight dataset. Apparently, after grouping the inputs, the model can better extract the temporal information from between frames and thus better recover the background.

\subsection{Comparison with State-of-the-Art}

We compare it with some state-of-the-art video rain removal methods and single-image rain removal methods, including MPRNet~\cite{zamir2021multi}, VRGNet~\cite{wang2021rain}, FastDeRain ~\cite{jiang2019fastderain}, JORDER~\cite{yang2017deep},  J4R-Net~\cite{liu2018erase}, SpacCNN~\cite{chen2018robust}, DualFlow~\cite{yang2019frame}, TMICS~\cite{mu2021triple}, ESTINet~\cite{zhang2022enhanced}, RMFD~\cite{yang2021recurrent} and MFDNN~\cite{su2022complex}. 

\textbf{Comparing on different datasets.} We show in Table~\ref{table:comparison sota} the derain results for all methods on the three datasets. Our approach uses only one model to deal with different types of rain. In this case, the proposed method still achieves good results. It can be seen that our method is significantly better than other SOTA methods. Compare to the latest and best method MFDNN, our method achieves gains of 1.32dB, 0.81dB, and 2.28dB in PSNR on the RainSynLight25, RainSynComplex25, and NTU-Rain datasets respectively. The above results demonstrate that our method is more effective in removing the different types of rain scenes.

\textbf{Comparing on real-world video frames.} We further compare the performance difference between the proposed and SOTA methods on real videos. The top row and the bottom row respectively are the videos from the NTU dataset and "mixkit.co"\footnote{\href{mixkit.co}{mixkit.co}}. As shown in Fig.~\ref{fig:vision_real} that our method retains most of the background information while removing the rain streaks more cleanly.

\subsection{Efficiency Analysis}

Table~\ref{table:runtime} shows the running speed of different advanced methods. All methods are based on the PyTorch implementation. We test all SOTA methods uniformly on a Linux system with the GeForce GTX 2080 Ti GPU. The test video resolution is $832 \times 512$. As can be seen from Table~\ref{table:runtime}, the number of parameters in our method is comparable to other deep learning methods, but the running speed and derain effect are far superior to other methods.

\section{Conclusion}
In this work, to avoid catastrophic forgetting, we design a rain review-based general video derain network via knowledge distillation. The method uses an old task model to guide the current model in learning new rain streaks knowledge thus avoiding forgetting. We also design a frame grouping encoder-decoder network, thus making full use of the temporal information of the video. Extensive experiments demonstrate that our proposed method outperforms state-of-the-art methods in terms of derain performance and running time.

\section*{Acknowledgment}
This work is supported by Natural Science Foundation of China (Grant No. 62202429, U20A20196, 61976191), Zhejiang Provincial Natural Science Foundation of China (Grant No. LY23F020024, LR21F020002 and LY23F020023) and the Hangzhou AI major scientific and technological innovation project (Grant No. 2022AIZD0061).


\bibliographystyle{IEEEtran}
\bibliography{reference}

\end{document}